\documentclass[11pt,a4paper]{article}
\pdfoutput=1
\usepackage[hyperref]{emnlp2020}
\usepackage{times}
\usepackage{latexsym}

\usepackage{microtype}

\usepackage{multirow}
\usepackage{hhline}
\newcommand{\bold}{\fontseries{b}\selectfont}

\aclfinalcopy 
\def\aclpaperid{1534} 

\newcommand{\comment}[1]{}

\newcommand*\samethanks[1][\value{footnote}]{\footnotemark[#1]}

\title{Multiple Word Embeddings for Increased Diversity of Representation}

\author{Brian Lester\thanks{\enspace Equal contribution; authors listed alphabetically}, Daniel Pressel\samethanks, Amy Hemmeter,\\
{\bf Sagnik Ray Choudhury and Srinivas Bangalore} \\
  \\ Interactions, Ann Arbor MI 48104 \\
  {\tt \{blester, dpressel, ahemmeter, schoudhury, sbangalore\}} \\
  {\tt @interactions.com} \\}

\date{}

\begin{document}
\maketitle
\begin{abstract}
Most state-of-the-art models in natural language processing (NLP) are neural models built on top
of large, pre-trained, contextual language models that generate representations of words in
context and are fine-tuned for the task at hand. The improvements afforded by these ``contextual
embeddings'' come with a high computational cost. In this work, we explore a simple technique that
substantially and consistently improves performance over a strong baseline with negligible increase in run time.
We concatenate multiple pre-trained embeddings to strengthen our representation of
words. We show that this concatenation technique works across many tasks, datasets, and model types.
We analyze aspects of pre-trained embedding similarity and vocabulary coverage and find that the
representational diversity between different pre-trained embeddings is the driving force of why this
technique works. We provide open source implementations of our models in both TensorFlow and
PyTorch.
\end{abstract}

\section{Introduction}

Much of the recent work in NLP has focused on better feature representations via contextual word embeddings
\cite{Peters:2018, Peters2017SemisupervisedST:17, RadfordTransformer2018, Akbik2018ContextualSE, devlin2018bert}.
These models vary in architecture and pre-training objective but they all encode the input based on the
surrounding context in some way. These papers normally compare to baselines like a bidirectional LSTM-CRF (biLSTM-CRF)
where words are represented by a \textit{single} pre-trained word embedding.

\citet{Peters:2018, Peters2017SemisupervisedST:17} and \citet{Akbik2018ContextualSE} pre-train large language models
based on LSTMs. Task-specific architectures are then built on top of these pre-trained models. \citet{Peters:2018}
introduce a technique for extracting word representations as a linear combination of layers in the pre-trained model.
Gradient updates are only applied to this weighting factor, which simplifies the training to some extent, but forward
propagation is still required for the full network which makes the model slow to train and evaluate.

\citet{RadfordTransformer2018}, followed by \citet{devlin2018bert}, pre-train deep transformers \cite{Vaswani:2017} on
massive corpora. They both use a simple output layer on top of the pre-trained model and tune the parameters of the
whole model. In this case,
training requires the forward and backward pass of the entire pre-trained model, which
has a significant impact on size and speed. \citet{devlin2018bert} used
specialized hardware which may be unrealistic for many inference scenarios.

The prevailing wisdom is that, because these pre-trained models are contextual, they can create representations
of a word that is different in different contexts.  For example, a polysemous word can be represented by different
vectors when its context suggests a different sense of a word, while context-independent word vectors need to represent a mix
of all the senses of a word. The majority of NLP models have a similar ``contextualization'' step, typically done via a
biLSTM, convolutional layers, or self-attention, but it is only learned from a smaller, task-specific corpus in contrast
to the massive corpora used by contextual embeddings.

Contextual embeddings and transfer learning architectures are slow to train and evaluate, which may make them infeasible
for many types of deployments. Using multiple pre-trained embeddings trained on different datasets, we can exploit the
bias in different datasets that results in different representations of the same word. By combining these embeddings, we
can create richer representations of the word without the high computational overhead required by contextual alternatives.
We find that the concatenation of multiple pre-trained word embeddings show
consistent improvements over single embeddings yielding results much closer to contextual alternatives.

\section{Experiments \& Results}

\begin{table*}[t]
\centering
\begin{tabular}{l l l l | r r r r}
    Task & Dataset & Model & Embeddings & mean & std & min & max \\
    \hhline{====|====}
    NER & CoNLL & biLSTM-CRF & 6B & 91.12 & 0.21 & 90.62 & 91.37 \\
    & & & Senna & 90.48 & 0.27 & 90.02 & 90.81 \\
    & & & 6B, Senna & \bold{91.61} & 0.25 & 91.15 & \bold{92.00} \\
    \cline{2-8}
    & WNUT-17 & biLSTM-CRF & 27B & 39.20 & 0.71 & 37.98 & 40.33 \\
    & & & 27B, w2v-30M & 39.52 & 0.83 & 38.09 & 40.39 \\
    & & & 27B, w2v-30M, 840B & \bold{40.33} & 1.13 & 38.38 & \bold{41.99} \\
    \cline{2-8}
    & OntoNotes & biLSTM-CRF & 6B & 87.02 & 9.15 & 86.75 & 87.24 \\
    & & & 6B, Senna & \bold{87.41} & 0.16 & 87.14 & \bold{87.74} \\
    \hline
    Slot Filling & Snips & biLSTM-CRF & 6B & 95.84 & 0.29 & 95.39 & 96.21 \\
    & & & GN & 95.28 & 0.41 & 94.51 & 95.81 \\
    & & & 6B, GN & \bold{96.04} & 0.28 & 95.39 & \bold{96.35} \\ 
    \hline
    POS & TW-POS & biLSTM-CRF & w2v-30M & 89.21 & 0.28 & 88.72 & 89.74 \\
    & & & 27B & 89.63 & 0.19 & 89.35 & 89.92 \\
    & & & 27B, w2v-30M & 90.35 & 0.20 & 89.99 & 90.60 \\
    & & & 27B, w2v-30M, 840B & \bold{90.75} & 0.14 & 90.53 & \bold{91.02} \\
    \hline
    Classification & SST2 & LSTM & 840B & 88.39 & 0.45 & 87.42 & 89.07 \\
    & & & GN & 87.58 & 0.54 & 86.16 & 88.19 \\
    & & & 840B, GN & \bold{88.57} & 0.44 & 87.59 & \bold{89.24} \\
    \cline{2-8}
    & AG-NEWS & LSTM & 840B & 92.53 & 0.45 & 87.42 & 89.07 \\
    & & & GN & 92.20 & 0.18 & 91.80 & 92.40 \\
    & & & 840B, GN & \bold{92.60} & 0.20 & 92.30 & \bold{92.86} \\
    \cline{2-8}
    & Snips & Conv & 840B &  97.47 & 0.33 & 97.01 & 97.86 \\
    & & & GN & 97.40 & 0.27 & 97.00 & 97.86 \\
    & & & 840B, GN & \bold{97.63} & 0.52 & 97.00 & \bold{98.29} \\
\end{tabular}
\caption{Results using multiple embeddings applied to several tasks and datasets. NER and Slot Filling tasks report
entity-level F1. POS tagging and Classification report token-level and example-level accuracy respectively. Using
multiple pre-trained embeddings helps across a wide range of tasks and datasets as well as across different model
architectures within a given task. All results are reported across 10 runs.} 
\label{tab:emb-results}
\end{table*}

We  use  three  sequential  prediction  tasks to test the performance of our concatenated embeddings: NER
(CoNLL 2003 \cite{TjongKimSang:2003:ICS:1119176.1119195}, WNUT-17 \cite{Derczynski2017ResultsOT}, and
OntoNotes \cite{Hovy:2006:O9S:1614049.1614064}), Slot filling (Snips \cite{Coucke2018SnipsVP}) and POS tagging
(TW-POS \cite{Gimpel:2011:PTT:2002736.2002747}). We also show results on three classification datasets:
SST2 \cite{D13-1170}, Snips intent classification \cite{Coucke2018SnipsVP}, and
AG-News\footnote{\label{foot:ag}\url{http://www.di.unipi.it/~gulli/AG_corpus_of_news_articles.html}}. For each
(task, dataset) pair we use the most common embedding used in literature, for example, GloVe embeddings were used for CONLL 2003 in \cite{ma-hovy:2016:P16-1:16} and Senna embeddings in \cite{Chiu2016NamedER,Peters:2018}. Embeddings were also chosen based on how well the embedding training data fit the task, i.e., we used GloVe vectors trained on twitter for the twitter part of speech tagging task. Once we developed tests for which embeddings worked together in Section \ref{sec:analysis} we checked if there were any more embeddings combinations we should try but did not find any additional combinations. For all tagging tasks, a biLSTM-CRF
model with convolutional character compositional inputs, following \cite{ma-hovy:2016:P16-1:16}, is used. For all classification tasks, a single layer LSTM model is used except for the Snips classification
dataset, where a convolutional word-based model \cite{kim-2014-convolutional} is used. The hyperparameters are
omitted here for brevity but can be found in our implementation.

The results are presented in Table \ref{tab:emb-results}. 6B, 27B and 840B are well-known, pre-trained GloVe
embeddings \cite{pennington2014glove} distributed via the authors site, w2v-30M \cite{Baseline:2018} and
GN \cite{mikolovgn} are Word2Vec embeddings trained on a corpus of 30 million tweets and Google News respectively,
and the Senna embeddings were trained by \citet{journals/jmlr/CollobertWBKKK11:11}.

We leverage multiple pre-trained embeddings in a model by creating one embeddings table per pre-trained embedding. Each input token in embedded into each vector space and the resulting vectors are concatenated into a single vector. This means that it is possible for there to be a type that is unattested in one pre-trained embedding vocabulary but present in the other. This results in a pre-trained vector from one embedding being concatenated with a randomly initialized vector form the other embedding space.

As hypothesized, we see improvements across tasks, datasets, and model architectures when using multiple embeddings.

Models using the concatenation of pre-trained and randomly initialized embeddings do $0.6$\% worse on average
compared to models that only use a single pre-trained embedding. This demonstrates that the performance
gains are from the combination of different pre-trained embeddings rather than the increase in the number of
parameters in the model. In some cases we were able to improve results further by adding several sets of
additional embeddings.

\begin{table}[h]
\centering
\begin{tabular}{l l | r }
    Task &  Domain & $\Delta$ \\
    \hhline{==|=}
    NER & General NER & 0.51 \\
    \hline
    Slot Filling & Automotive & 0.14 \\
                 & Cyber Security & 0.06 \\
                 & Customer Service & 0.34 \\
    \hline
    Intent & Automotive & 0.52 \\
           & Cyber Security & 0.03 \\
           & Customer Service & 0.16 \\
\end{tabular}
\caption{
    Performance using multiple embeddings on internal datasets. Although smaller than well-known datasets,
    we see consistent improvements across internal tasks and domains.
}
\label{tab:internal-results}
\end{table}

\begin{table}[h]
\centering
\begin{tabular}{l l | r}
    Task & Dataset & $\Delta$ \\
    \hhline{==|=}
    NER & CoNLL & 0.54 \\
        & WNUT-17 & 2.88 \\
        & OntoNotes & 0.45 \\
    \hline
    Slot Filling & Snips & 0.21 \\
    \hline
    POS & TW-POS & 1.25 \\
    \hline
    Classification & SST2 & 0.20 \\
                   & AG-NEWS & 0.08 \\
                   & Snips & 0.16 \\
\end{tabular}
\caption{Relative difference for well-known datasets to help frame the results in Table \ref{tab:internal-results}}
\label{tab:emb-relative}
\end{table}

Table \ref{tab:internal-results} summarizes the results of using the multiple embedding approach on internal datasets.
These datasets are drawn from the tasks defined earlier and span a variety of specialized domains. Due to the nature
of the datasets the results are presented as the relative change in performance. Table \ref{tab:emb-relative} is provided to help frame the relative performance numbers from the internal datasets.
 
The models were trained with MEAD/Baseline \cite{Baseline:2018}, an open-source framework for
developing, training, and deploying NLP models.
 
\section{Analysis}
\label{sec:analysis}

\begin{table*}[t]
    \centering
    \begin{tabular}{l l | r r r}
    Dataset & Embeddings & mean & std & max \\
    \hhline{==|===}
    SST2 & GN & 87.58 & 0.54 & 88.19 \\
         & GN + Random init & 87.62 & 0.22 & 88.64 \\
         & GN + 840B complement to GN & 87.72 & 0.23 & 88.02 \\
         & GN + 840B matched to GN & 88.53 & 0.55 & 89.45 \\
         & GN + 840B & 88.57 & 0.44 & 89.24 \\
    \hline
    CoNLL & 6B & 91.12 & 0.21 & 91.37 \\
          & 6B + Random init & 90.77 & 0.17 & 91.11 \\
          & 6B + Senna complement to 6B & 90.73 & 0.29 & 91.19 \\
          & 6B + Senna matched to 6B & 91.47 & 0.18 & 91.78 \\
          & 6B + Senna & 91.61 & 0.25 & 92.00
    \end{tabular}
    \caption{
        An ablation to explain why multiple embeddings work. The majority of the improvement comes the case where
        we take only the words from the second pre-trained embedding that appear in the first vocab (the matched row).
        This suggests that having different representations for a word is much more important than increased model
        capacity (tested in the Random init row) or the increased coverage in the pre-trained vocabulary (represented
        by the complement row).
    }
    \label{tab:why-emb}
\end{table*}

\begin{table*}[t]
    \centering
    \begin{tabular}{l | c c c c c c}
                   & \multicolumn{2}{c}{Overlap} & \multicolumn{2}{c}{Attested} & \multicolumn{2}{c}{Performance} \\
        Embeddings & train & dev & train & dev & mean & std \\
        \hhline{=|======}
        Senna & 18.9 & 20.8 & 74.3 & 80.3 & 91.610 & 0.247 \\
        GloVe twitter 27B & 24.9 & 27.2 & 68.1 & 76.1 & 91.098 & 0.135 \\
        GloVe 840B & 41.7 & 40.6 & 83.2 & 88.5 & 91.011 & 0.228 \\
        GloVe 42B & 45.5 & 45.3 & 90.4 & 93.8 & 91.163 & 0.146 \\
        GoogleNews & 25.2 & 26.8 & 55.9 & 65.1 & 90.948 & 0.180 \\

    \end{tabular}
    \caption{
        Embedding similarity as defined by average Jaccard similarity of the 10 nearest neighbors on the top
        200 words in CoNLL 2003. Performance is the entity-level F1 score of each embedding when paired with
        Glove 6B 100 dimension embeddings. Here we can see that using pairs of dissimilar embeddings correlate
        with better performance as long as the embeddings have enough coverage to be effectively leveraged.
    }
    \label{tab:overlap}
\end{table*}

There are three logical places where the observed improvements could come from. 1) The use of multiple pre-trained embeddings
creates a slightly
larger model, increasing the network capacity---the embeddings are larger and therefore the projection from the embeddings
to the first layer of the model will also be slightly bigger. 2) The use of a second pre-trained embedding increases the 
vocabulary size and more words are attested.  A word that has a pre-trained representation will start the model in a better spot than a randomly
initialized representation. 3) The second set of pre-trained embeddings gives a different perspective of the words. Most
pre-trained embeddings are trained on different data and encode different biases and senses into the embedding
that reflect the quirks and unique contexts found in the pre-training data. This representational diversity
will allow a model to capitalize on different senses, or the combination of senses, that would not be present when using a single embedding.

In order to tease apart which of these factors are at play we designed a series of models that aim to isolate
each effect and report results in Table \ref{tab:why-emb}. First, we train a model that uses a single pre-trained embedding
and a second set of vectors that are initialized randomly. If the main improvement is due to increased model capacity this
configuration should perform well. The second model uses a special version of the second pre-trained embedding where we
remove all the words that already appear in the original pre-trained vocabulary. In this second set of embeddings, randomly
initialized vectors are used for the words that are covered in the original vocabulary in order to keep the embeddings size
consistent with the previous model. If the main reason for improvement is the increased vocabulary coverage, this model should perform well. The final version
of the model also uses a customized version of the second pre-trained embedding. This time we only keep embeddings that are already represented in the original vocabulary. This is designed to test if the main source of improvement is the difference in the representations each pre-trained embedding brings to the table.

From our ablation studies using the above variations on both the SST2 and CoNLL datasets and find that the most
important thing is the representational diversity in the pre-trained embeddings. This dovetails nicely with our observation that
embeddings trained on distinct datasets tend to perform well together. To further test this hypothesis, we look at the ``similarity'' of various pre-trained embeddings. We define ``similarity'' using the overlap of nearest neighbors in the embedding space as in \citet{wendlandt-etal-2018-factors}. Specificly we use the average Jaccard overlap percentage between the 10 nearest neighbors for each of the top 200 words in the dataset by frequency. Table \ref{tab:overlap} shows the overlap of different embeddings with the Glove 6B 100 dimension embedding and how their combination affects the performance. As it can be seen, Senna has the lowest overlap and causes the biggest performance gain. 

However, this does not hold for the GoogleNews embedding which also has a low overlap yet the combination actually causes a drop in performance. This can be explained by coverage---the percentage of unique types in the data that are attested in the pre-trained vocabulary. That number is surprisingly low for GoogleNews and causes the GoogleNews representations to be used so rarely they actually cause a drop in performance.

In summary, one should look for two characteristics when combining embeddings: the word representations should have low ``similarity" and the unique types in the dataset should be highly attested in both pre-trained vocabularies.

\section{Conclusion}

Recent large-scale, contextual, pre-trained models are exciting but produce relatively slow models.
We propose a simple, lightweight technique: concatenation of pre-trained embeddings. We show that this technique has a significant
impact on error reduction and a negligible effect of speed.

However, the concatenation on any two random pre-trained embeddings is not guaranteed to work well. From our analysis, we are able to suggest a recipe for finding an effective combination: there should be a high degree of coverage  of the unique types in each of the pre-trained embedding vocabularies and the word vectors should exhibit representational diversity. In future work, we intend to try other methods of embeddings combination while remaining computationally cheap. We also plan to find more principled ways to quantify the diversity in pre-trained embeddings, which can suggest ways to induce representational diversity into the embedding pre-training procedure itself.


\bibliography{emnlp2020}

\begin{thebibliography}{20}
\expandafter\ifx\csname natexlab\endcsname\relax\def\natexlab#1{#1}\fi

\bibitem[{Akbik et~al.(2018)Akbik, Blythe, and
  Vollgraf}]{Akbik2018ContextualSE}
Alan Akbik, Duncan Blythe, and Roland Vollgraf. 2018.
\newblock \href {https://www.aclweb.org/anthology/C18-1139} {{C}ontextual
  {S}tring {E}mbeddings for {S}equence {L}abeling}.
\newblock In \emph{Proceedings of the 27th International Conference on
  Computational Linguistics}, pages 1638--1649, Santa Fe, New Mexico, USA.
  Association for Computational Linguistics.

\bibitem[{Chiu and Nichols(2016)}]{Chiu2016NamedER}
Jason~P.C. Chiu and Eric Nichols. 2016.
\newblock \href {https://doi.org/10.1162/tacl_a_00104} {{N}amed {E}ntity
  {R}ecognition with {B}idirectional {LSTM}-{CNN}s}.
\newblock \emph{Transactions of the Association for Computational Linguistics},
  4:357--370.

\bibitem[{Collobert et~al.(2011)Collobert, Weston, Bottou, Karlen, Kavukcuoglu,
  and Kuksa}]{journals/jmlr/CollobertWBKKK11:11}
Ronan Collobert, Jason Weston, Léon Bottou, Michael Karlen, Koray Kavukcuoglu,
  and Pavel~P. Kuksa. 2011.
\newblock \href
  {http://dblp.uni-trier.de/db/journals/jmlr/jmlr12.html#CollobertWBKKK11}
  {{N}atural {L}anguage {P}rocessing ({A}lmost) from {S}cratch.}
\newblock \emph{Journal of Machine Learning Research}, 12:2493--2537.

\bibitem[{Coucke et~al.(2018)Coucke, Saade, Ball, Bluche, Caulier, Leroy,
  Doumouro, Gisselbrecht, Caltagirone, Lavril, Primet, and
  Dureau}]{Coucke2018SnipsVP}
Alice Coucke, Alaa Saade, Adrien Ball, Th{\'e}odore Bluche, Alexandre Caulier,
  David Leroy, Cl{\'e}ment Doumouro, Thibault Gisselbrecht, Francesco
  Caltagirone, Thibaut Lavril, Ma{\"e}l Primet, and Joseph Dureau. 2018.
\newblock {S}nips {V}oice {P}latform: an {E}mbedded {S}poken {L}anguage
  {U}nderstanding {S}ystem for {P}rivate-by-design {V}oice {I}nterfaces.
\newblock \emph{arXiv preprint}, arXiv:1805.10190.

\bibitem[{Derczynski et~al.(2017)Derczynski, Nichols, van Erp, and
  Limsopatham}]{Derczynski2017ResultsOT}
Leon Derczynski, Eric Nichols, Marieke van Erp, and Nut Limsopatham. 2017.
\newblock \href {https://doi.org/10.18653/v1/W17-4418} {{R}esults of the
  {WNUT}2017 {S}hared {T}ask on {N}ovel and {E}merging {E}ntity {R}ecognition}.
\newblock In \emph{Proceedings of the 3rd Workshop on Noisy User-generated
  Text}, pages 140--147, Copenhagen, Denmark. Association for Computational
  Linguistics.

\bibitem[{Devlin et~al.(2019)Devlin, Chang, Lee, and
  Toutanova}]{devlin2018bert}
Jacob Devlin, Ming-Wei Chang, Kenton Lee, and Kristina Toutanova. 2019.
\newblock \href {https://doi.org/10.18653/v1/N19-1423} {{BERT}: {P}re-training
  of {D}eep {B}idirectional {T}ransformers for {L}anguage {U}nderstanding}.
\newblock pages 4171--4186.

\bibitem[{Gimpel et~al.(2011)Gimpel, Schneider, O'Connor, Das, Mills,
  Eisenstein, Heilman, Yogatama, Flanigan, and
  Smith}]{Gimpel:2011:PTT:2002736.2002747}
Kevin Gimpel, Nathan Schneider, Brendan O'Connor, Dipanjan Das, Daniel Mills,
  Jacob Eisenstein, Michael Heilman, Dani Yogatama, Jeffrey Flanigan, and
  Noah~A. Smith. 2011.
\newblock \href {http://dl.acm.org/citation.cfm?id=2002736.2002747}
  {Part-of-speech tagging for twitter: annotation, features, and experiments}.
\newblock In \emph{Proceedings of the 49th Annual Meeting of the Association
  for Computational Linguistics: Human Language Technologies: short papers -
  Volume 2}, HLT '11, pages 42--47, Stroudsburg, PA, USA. Association for
  Computational Linguistics.

\bibitem[{Hovy et~al.(2006)Hovy, Marcus, Palmer, Ramshaw, and
  Weischedel}]{Hovy:2006:O9S:1614049.1614064}
Eduard Hovy, Mitchell Marcus, Martha Palmer, Lance Ramshaw, and Ralph
  Weischedel. 2006.
\newblock \href {http://dl.acm.org/citation.cfm?id=1614049.1614064}
  {{O}nto{N}otes: {T}he 90\% {S}olution}.
\newblock In \emph{Proceedings of the Human Language Technology Conference of
  the NAACL, Companion Volume: Short Papers}, NAACL-Short '06, pages 57--60,
  Stroudsburg, PA, USA. Association for Computational Linguistics.

\bibitem[{Kim(2014)}]{kim-2014-convolutional}
Yoon Kim. 2014.
\newblock \href {https://doi.org/10.3115/v1/D14-1181} {{C}onvolutional {N}eural
  {N}etworks for {S}entence {C}lassification}.
\newblock In \emph{Proceedings of the 2014 Conference on Empirical Methods in
  Natural Language Processing ({EMNLP})}, pages 1746--1751, Doha, Qatar.
  Association for Computational Linguistics.

\bibitem[{Ma and Hovy(2016)}]{ma-hovy:2016:P16-1:16}
Xuezhe Ma and Eduard Hovy. 2016.
\newblock \href {http://www.aclweb.org/anthology/P16-1101} {{E}nd-to-end
  {S}equence {L}abeling via {B}i-directional {LSTM}-{CNN}s-{CRF}}.
\newblock In \emph{Proceedings of the 54th Annual Meeting of the Association
  for Computational Linguistics (Volume 1: Long Papers)}, pages 1064--1074,
  Berlin, Germany. Association for Computational Linguistics.

\bibitem[{Mikolov et~al.(2013)Mikolov, Chen, Corrado, and Dean}]{mikolovgn}
Tomas Mikolov, Kai Chen, Greg~S. Corrado, and Jeffrey Dean. 2013.
\newblock \href {http://arxiv.org/abs/1301.3781} {{E}fficient {E}stimation of
  {W}ord {R}epresentations in {V}ector {S}pace}.

\bibitem[{Pennington et~al.(2014)Pennington, Socher, and
  Manning}]{pennington2014glove}
Jeffrey Pennington, Richard Socher, and Christopher~D. Manning. 2014.
\newblock \href {http://www.aclweb.org/anthology/D14-1162} {{G}lo{V}e: {G}lobal
  {V}ectors for {W}ord {R}epresentation}.
\newblock In \emph{Empirical Methods in Natural Language Processing (EMNLP)},
  pages 1532--1543.

\bibitem[{Peters et~al.(2018)Peters, Neumann, Iyyer, Gardner, Clark, Lee, and
  Zettlemoyer}]{Peters:2018}
Matthew Peters, Mark Neumann, Mohit Iyyer, Matt Gardner, Christopher Clark,
  Kenton Lee, and Luke Zettlemoyer. 2018.
\newblock \href {https://doi.org/10.18653/v1/N18-1202} {{D}eep {C}ontextualized
  {W}ord {R}epresentations}.
\newblock In \emph{Proceedings of the 2018 Conference of the North {A}merican
  Chapter of the Association for Computational Linguistics: Human Language
  Technologies, Volume 1 (Long Papers)}, pages 2227--2237, New Orleans,
  Louisiana. Association for Computational Linguistics.

\bibitem[{Peters et~al.(2017)Peters, Ammar, Bhagavatula, and
  Power}]{Peters2017SemisupervisedST:17}
Matthew~E. Peters, Waleed Ammar, Chandra Bhagavatula, and Russell Power. 2017.
\newblock \href {https://doi.org/10.18653/v1/P17-1161} {{S}emi-supervised
  {S}equence {T}agging with {B}idirectional {L}anguage {M}odels}.
\newblock In \emph{Proceedings of the 55th Annual Meeting of the Association
  for Computational Linguistics, {ACL} 2017, Vancouver, Canada, July 30 -
  August 4, Volume 1: Long Papers}, pages 1756--1765.

\bibitem[{Pressel et~al.(2018)Pressel, Ray~Choudhury, Lester, Zhao, and
  Barta}]{Baseline:2018}
Daniel Pressel, Sagnik Ray~Choudhury, Brian Lester, Yanjie Zhao, and Matt
  Barta. 2018.
\newblock \href {http://aclweb.org/anthology/W18-2506} {{B}aseline: {A}
  {L}ibrary for {R}apid {M}odeling, {E}xperimentation and {D}evelopment of
  {D}eep {L}earning {A}lgorithms {T}argeting {NLP}}.
\newblock In \emph{Proceedings of Workshop for NLP Open Source Software
  (NLP-OSS)}, pages 34--40. Association for Computational Linguistics.

\bibitem[{Radford et~al.(2018)Radford, Narasimhan, Salimans, and
  Sutskever}]{RadfordTransformer2018}
Alec Radford, Karthik Narasimhan, Tim Salimans, and Ilya Sutskever. 2018.
\newblock \href
  {https://s3-us-west-2.amazonaws.com/openai-assets/research-covers/language-unsupervised/language_understanding_paper.pdf}
  {{I}mproving {L}anguage {U}nderstanding by {G}enerative {P}re-{T}raining}.

\bibitem[{Socher et~al.(2013)Socher, Perelygin, Wu, Chuang, Manning, Ng, and
  Potts}]{D13-1170}
Richard Socher, Alex Perelygin, Jean Wu, Jason Chuang, Christopher~D. Manning,
  Andrew Ng, and Christopher Potts. 2013.
\newblock \href {http://aclweb.org/anthology/D13-1170} {{R}ecursive {D}eep
  {M}odels for {S}emantic {C}ompositionality {O}ver a {S}entiment {T}reebank}.
\newblock In \emph{Proceedings of the 2013 Conference on Empirical Methods in
  Natural Language Processing}, pages 1631--1642. Association for Computational
  Linguistics.

\bibitem[{Tjong Kim~Sang and
  De~Meulder(2003)}]{TjongKimSang:2003:ICS:1119176.1119195}
Erik~F. Tjong Kim~Sang and Fien De~Meulder. 2003.
\newblock \href {https://doi.org/10.3115/1119176.1119195} {{I}ntroduction to
  the {C}o{NLL}-2003 {S}hared {T}ask: {L}anguage-independent {N}amed {E}ntity
  {R}ecognition}.
\newblock In \emph{Proceedings of the Seventh Conference on Natural Language
  Learning at HLT-NAACL 2003 - Volume 4}, CONLL '03, pages 142--147,
  Stroudsburg, PA, USA. Association for Computational Linguistics.

\bibitem[{Vaswani et~al.(2017)Vaswani, Shazeer, Parmar, Uszkoreit, Jones,
  Gomez, Kaiser, and Polosukhin}]{Vaswani:2017}
Ashish Vaswani, Noam Shazeer, Niki Parmar, Jakob Uszkoreit, Llion Jones,
  Aidan~N Gomez, \L~ukasz Kaiser, and Illia Polosukhin. 2017.
\newblock \href
  {http://papers.nips.cc/paper/7181-attention-is-all-you-need.pdf} {{A}ttention
  is {A}ll {Y}ou {N}eed}.
\newblock In I.~Guyon, U.~V. Luxburg, S.~Bengio, H.~Wallach, R.~Fergus,
  S.~Vishwanathan, and R.~Garnett, editors, \emph{Advances in Neural
  Information Processing Systems 30}, pages 5998--6008. Curran Associates, Inc.

\bibitem[{Wendlandt et~al.(2018)Wendlandt, Kummerfeld, and
  Mihalcea}]{wendlandt-etal-2018-factors}
Laura Wendlandt, Jonathan~K. Kummerfeld, and Rada Mihalcea. 2018.
\newblock \href {https://doi.org/10.18653/v1/N18-1190} {{F}actors {I}nfluencing
  the {S}urprising {I}nstability of {W}ord {E}mbeddings}.
\newblock In \emph{Proceedings of the 2018 Conference of the North {A}merican
  Chapter of the Association for Computational Linguistics: Human Language
  Technologies, Volume 1 (Long Papers)}, pages 2092--2102, New Orleans,
  Louisiana. Association for Computational Linguistics.

\end{thebibliography}
\bibliographystyle{acl_natbib}

\appendix

\section{Reproducibility}

\subsection{Hyperparameters}

Mead/Baseline is a configuration file driven model training framework. All hyperparameters are fully specified in
the congifuration files included with the source code for our experiments.

\subsection{Computational Resources}

All models were trained on a single NVIDIA 1080Ti. While multiple GPUs were used for training many models in parallel
to facilitate a testing many datasets and to estimate the variability of the method the actual model can easily be trained
on a single GPU.

\subsection{Evaluation}

To calculate metrics, entity-level F1 is used for NER and slot-filling. In entity level F1 first
entities are created from the token level labels and compared to the gold ones. Entities that
match on both type and boundaries are considered correct while a mismatch in either causes an
error. The F1 score is then calculated from these entities. Accuracy is used for classification and
part of speech tagging. Accuracy is defined as the proportion of correct elements to all
elements. In classification a single example is an element. In part of speech tagging each token is
an element so our accuracy is the the number of correct tokens divided by the number of tokens in the dataset.
We use the evaluation code that ships with the framework we use, MEAD/Baseline, which we have bundled
with the source code of our experiments.

\begin{table*}[ht]
\centering
\begin{tabular}{l l l l | r }
    Task & Dataset & Model & Embeddings & Number of parameters \\
    \hhline{====|=}
    NER & CoNLL & biLSTM-CRF & 6B & 3,234,440  \\
    & & & Senna & 1,810,690  \\
    & & & 6B, Senna & 4,658,190 \\
    \cline{2-5}
    & WNUT-17 & biLSTM-CRF & 27B & 3,849,632 \\
    & & & 27B, w2v-30M & 6,499,532  \\
    & & & 27B, w2v-30M, 840B & 12,090,032 \\
    \cline{2-5}
    & OntoNotes & biLSTM-CRF & 6B & 5,569,382 \\
    & & & 6B, Senna & 7,673,632  \\
    \hline
    Slot Filling & Snips & biLSTM-CRF & 6B & 1,819,466 \\
    & & & GN & 4,567,066 \\
    & & & 6B, GN & 5,940,866 \\ 
    \hline
    POS & TW-POS & biLSTM-CRF & w2v-30M & 1,241,332 \\
    & & & 27B & 1,788,982 \\
    & & & 27B, w2v-30M & 2,908,132 \\
    & & & 27B, w2v-30M, 840B & 5,408,332 \\
    \hline
    Classification & SST2 & LSTM & 840B & 6,456,702 \\
    & & & GN & 6,456,702 \\
    & & & 840B, GN & 12,109,002 \\
    \cline{2-5}
    & AG-NEWS & LSTM & 840B & 20,842,604 \\
    & & & GN & 20,842,604 \\
    & & & 840B, GN & 41,522,804 \\
    \cline{2-5}
    & Snips & Conv & 840B & 4,003,807 \\
    & & & GN & 4,003,807 \\
    & & & 840B, GN & 8,005,207 \\
\end{tabular}
\caption{The number of parameters for different models.} 
\label{tab:parameter-counts}
\end{table*}

\subsection{Dataset Information}

\begin{table*}[ht]
    \centering
    \begin{tabular}{l l | r r r r}
        Dataset & & Train & Dev & Test & Total \\
        \hhline{==|====}
        CoNLL & Examples & 14,987 & 3,466 & 3674 & 22137 \\
              & Tokens & 204,567 & 51,578 & 46,666 & 302,811 \\
        WNUT-17 & Examples & 3,394 & 1,009 & 1,287 & 5,690 \\
                & Tokens & 62,730 & 15,733 & 23,394 & 101,857 \\
        OntoNotes & Examples & 59,924 & 8,528 & 8,262 & 76,714 \\
                  & Tokens & 1,088,503 & 147,724 & 152,728 & 1,388,955 \\
        Snips & Examples & 13,084 & 700 & 700 & 14,484  \\
              & Tokens & 117,700 & 6,384 & 6,354 & 130,438 \\
        TW-POS & Examples & 1,000 & 327 & 500 & 1,827 \\
               & Tokens & 14,619 & 4,823 & 7,152 & 26,594 \\
        SST2 & Examples & 76,961 & 872 & 1,821 & 79,654 \\
             & Tokens & 717,127 & 17,046, & 35,023 & 769,196 \\
        AG-NEWS & Examples & 110,000 & 10,000 & 7,600 & 127,600 \\
                & tokens & 4,806,909 & 433,659 & 329,617 & 5,570,185
    \end{tabular}
    \caption{Example and token count statistics for public datasets used.}
    \label{tab:data-stats}
\end{table*}

Relevant information about datasets can be found in Table \ref{tab:data-stats}. The majority of data is used as distributed
except we convert NER and slot-filling datasets to the IOBES format. All public dataset used are included in the supplementary
material. A quick overview of each dataset follows:

\textbf{CoNLL}: A NER dataset based on news text. We converted the IOB labels into the IOBES format. There are 4 entity types, \texttt{MISC}, \texttt{LOC}, \texttt{PER}, and \texttt{LOC}.

\textbf{WNUT-17}: A NER dataset of new and emerging entities based on noisy user text. We converted the BIO labels into the IOBES format. There are 6 entity types, \texttt{corporation},
\texttt{creative-work}, \texttt{group}, \texttt{location}, \texttt{person}, and \texttt{product}.

\textbf{OntoNotes}: A much larger NER dataset. We converted the labels into the IOBES format. There are 18 entity types, \texttt{CARDINAL}, \texttt{DATE}, \texttt{EVENT}, \texttt{FAC}, \texttt{GPE}, \texttt{LANGUAGE}, \texttt{LAW}, \texttt{LOC}, \texttt{MONEY}, \texttt{NORP}, \texttt{ORDINAL}, \texttt{ORG}, \texttt{PERCENT}, \texttt{PERSON}, \texttt{PRODUCT}, \texttt{QUANTITY}, \texttt{TIME}, and \texttt{WORK\_OF\_ART}.

\textbf{Snips}: A slot-filling dataset focusing on commands one would give a virtual assistant. We converted the dataset from its normal format of two associated files, one containing surface terms and one containing labels to the more standard CoNLL file format and converted the labels to the IOBES format. There are 39 entity types, \texttt{album}, \texttt{artist}, \texttt{best\_rating}, \texttt{city}, \texttt{condition\_description}, \texttt{condition\_temperature}, \texttt{country}, \texttt{cuisine}, \texttt{current\_location}, \texttt{entity\_name}, \texttt{facility}, \texttt{genre}, \texttt{geographic\_poi}, \texttt{location\_name}, \texttt{movie\_name}, \texttt{movie\_type}, \texttt{music\_item}, \texttt{object\_location\_type}, \texttt{object\_name}, \texttt{object\_part\_of\_series\_type}, \texttt{object\_select}, \texttt{object\_type}, \texttt{party\_size\_description}, \texttt{party\_size\_number}, \texttt{playlist}, \texttt{playlist\_owner}, \texttt{poi}, \texttt{rating\_unit}, \texttt{rating\_value}, \texttt{restaurant\_name}, \texttt{restaurant\_type}, \texttt{served\_dish}, \texttt{service}, \texttt{sort}, \texttt{spatial\_relation}, \texttt{state}, \texttt{timeRange}, \texttt{track}, and \texttt{year}.

\textbf{TW-POS}: A twitter part of speech dataset. There are 25 parts of speech, \texttt{!}, \texttt{\#}, \texttt{\$}, \texttt{\&}, \texttt{,}, \texttt{@}, \texttt{A}, \texttt{D}, \texttt{E}, \texttt{G}, \texttt{L}, \texttt{M}, \texttt{N}, \texttt{O}, \texttt{P}, \texttt{R}, \texttt{S}, \texttt{T}, \texttt{U}, \texttt{V}, \texttt{X}, \texttt{Y}, \texttt{Z}, \texttt{\^}, and \texttt{\~}.

\textbf{SST2}: A binary sentiment analysis dataset based on movie reviews. We use the version where the training data is made up of phrases.

\textbf{AG-NEWS}: A four class text classification dataset for categorizing news data based on the 4 most common categories.
There is not a standardized train and development split (there is a defined test set) so we created our own split which is included in the
supplementary material.

\textbf{Snips-Intent}: The intent classification portion of the snips dataset. Again the intents pertain to requests one would
make to a virtual assitant. There are 7 intents, \texttt{SearchScreeningEvent}, \texttt{PlayMusic}, \texttt{AddToPlaylist}, \texttt{BookRestaurant}, \texttt{RateBook}, \texttt{SearchCreativeWork}, and \texttt{GetWeather}.

\end{document}